\documentclass{article}
\usepackage{pifont}
\usepackage{float}
\usepackage{subcaption}
\usepackage{booktabs} 
\usepackage[accepted]{arxiv}
\usepackage{eccvabbrv}
\usepackage{makecell}

\definecolor{purple}{RGB}{102,0,255}

\usepackage[pagebackref,breaklinks,colorlinks]{hyperref}
\hypersetup{
colorlinks=true,
linkcolor=red,
citecolor=purple,
urlcolor=black
}

\usepackage{balance}

\usepackage{mathtools, nccmath}
\DeclarePairedDelimiter{\nint}\lfloor\rceil

\papertitlerunning{TMPQ-DM}

\begin{document}

\twocolumn[
\papertitle{TMPQ-DM: Joint Timestep Reduction and Quantization Precision Selection for Efficient Diffusion Models}

\begin{paperauthorlist}
\paperauthor{Haojun Sun}{thu}
\paperauthor{Chen Tang}{thu}
\paperauthor{Zhi Wang}{thu}
\paperauthor{Yuan Meng}{thu}
\paperauthor{Jingyan Jiang}{sztu}
\paperauthor{Xinzhu Ma}{cuhk}
\paperauthor{Wenwu Zhu}{thu} \\
$^{1}$ Tsinghua University \quad $^{2}$ Shenzhen Technology University \quad $^{3}$ The Chinese University of Hong Kong
\end{paperauthorlist}

\paperaffiliation{thu}{Tsinghua University}
\paperaffiliation{cuhk}{The Chinese University of Hong Kong}
\paperaffiliation{sztu}{Shenzhen Technology University}

\vskip 0.3in

\begin{abstract}
Diffusion models have emerged as preeminent contenders in the realm of generative models, showcasing remarkable efficacy across diverse tasks and real-world scenarios. 
Distinguished by distinctive sequential generative processes of diffusion models, characterized by hundreds or even thousands of timesteps, diffusion models progressively reconstruct images from pure Gaussian noise, with each timestep necessitating full inference of the entire model. 
However, the substantial computational demands inherent to these models present challenges for deployment, quantization is thus widely used to lower the bit-width for reducing the storage and computing overheads. 
Nevertheless, current quantization methodologies primarily focus on model-side optimization, disregarding the temporal dimension, such as the length of the timestep sequence, thereby allowing redundant timesteps to continue consuming computational resources, leaving substantial scope for accelerating the generative process. 
In this paper, we introduce TMPQ-DM, which jointly optimizes timestep reduction and quantization to achieve a superior performance-efficiency trade-off, addressing both temporal and model optimization aspects. 
For timestep reduction, we devise a non-uniform grouping scheme tailored to the non-uniform nature of the denoising process, thereby mitigating the explosive combinations of timesteps. 
In terms of quantization, we adopt a fine-grained layer-wise approach to allocate varying bit-widths to different layers based on their respective contributions to the final generative performance, thus rectifying performance degradation observed in prior studies. 
To expedite the evaluation of fine-grained quantization, we further devise a super-network to serve as a precision solver by leveraging shared quantization results. 
These two design components are seamlessly integrated within our framework, enabling rapid joint exploration of the exponentially large decision space via a gradient-free evolutionary search algorithm. 
Experiments on 5 representative datasets demonstrate the effectiveness of proposed method, we achieve more than $10\times$ overall BitOPs savings while maintaining the comparable generative performance. 
Code will be available soon. 
\end{abstract}
]

\section{Introduction}
\label{sec:intro}

Diffusion models have recently attracted widespread attention, demonstrating significant potential across various generative tasks (\eg image generation, inpainting, and even object detection and segmentation), spanning multiple domains including image, text, speech, and biology ~\cite{liu2023diffusion,lugmayr2022repaint,chen2023diffusiondet,avrahami2022blended,bai2023dreamdiffusion}. Usually, diffusion models initiate with pure Gaussian noise and iteratively generate the desired image, achieving better generation quality than previous state-of-the-art methods like GANs~\cite{DiffusionBeatsGan, DiT, CADS}. Despite the excellent generation results, diffusion models require extensive generation steps, ranging from hundreds to thousands, and rely on complex neural networks, such as Transformers~\cite{AttentionIsAllYouNeed, bao2022one, DiT}, to parameterize the denoising process. This not only increases computational demands but also results in slower generation speeds compared to GAN models. For example, generating 50k 32$\times$32 images takes about 20 hours on a NVIDIA 2080 Ti GPU~\cite{DDIM}, whereas a GAN completes the task in under a minute.

Furthermore, the expansive parameter count of models like Stable Diffusion-XL \cite{SDXL}, which boasts 6.6 billion parameters, significantly burdens GPU memory, posing challenges for deployment in latency-sensitive and memory-constrained environments such as edge devices. To address these limitations, recent efforts have concentrated on developing efficient sampling methods~\cite{DDIM,dpm-solver,PLMS,DDSS} that aim to shorten the sampling trajectory. Nevertheless, these methods often overlook the substantial computational costs associated with each timestep due to the complexity of the underlying neural networks, leaving considerable room for improving the efficiency of denoising process~\cite{li2023q}.

Hence, there are also several attempts to compress the diffusion models for saving the single-step cost~\cite{li2023diffnas,li2023q,shang2023post,fang2024structural}, both adopting the model compression techniques such as neural architecture search~\cite{yu2020bignas,zoph2016neural}, pruning \cite{molchanov2019importance}, and quantization~\cite{tang2022arbitrary,zhou2016dorefa}. Among them, quantization is a promising method for lowering the runtime and storage overheads since its simplicity in avoiding architectural modifications. Specifically, post-training quantization (PTQ) is favored over quantization-aware training (QAT) as it does not require extensive fine-tuning on large datasets. However, existing studies neglect the meaningful interactions between timestep reduction (temporal optimization) and quantization (model optimization). For example, previous quantization methods mainly focus on transferring the quantization techniques from well-established research of discriminative models, neglecting the fact that some timesteps are rather redundant~\cite{li2023q,he2024ptqd,shang2023post}. For example, as shown in our experiment, we can generate comparable results with only 1.5\% timesteps on CIFAR-10 dataset.

\begin{figure*}[tb] 
  \centering
\includegraphics[width=\textwidth]{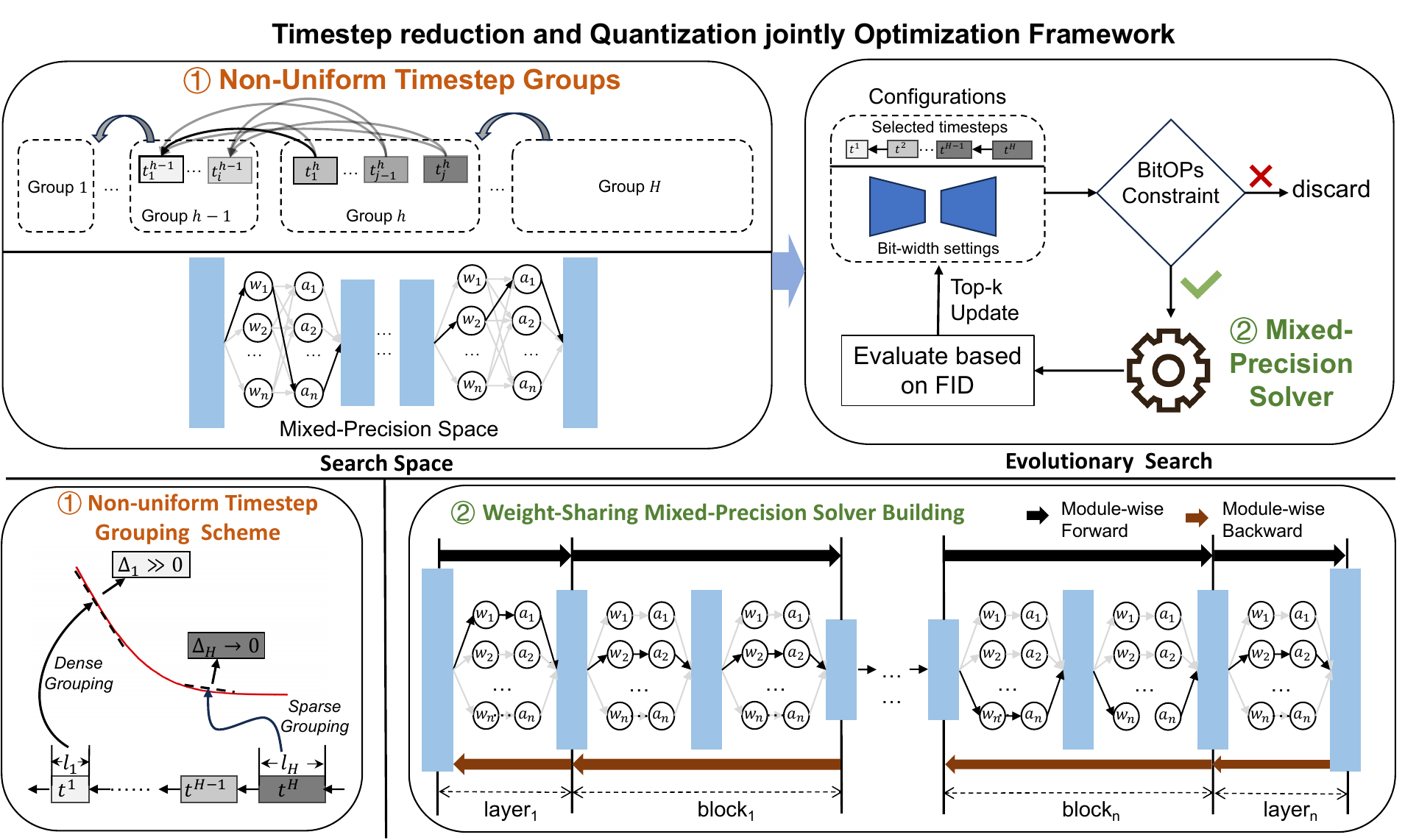}
\caption{
\emph{Top:} We jointly perform timestep sub-sequence search and fine-grained layer-wise quantization precision selection within a \emph{unified search space} via a gradient-free evolutionary search algorithm. 
\emph{Bottom:} $\textcircled{1}$ To deal with the explosive search space, we leverage a property of denoising process (Sec. \ref{sec:foundation}) to design a non-uniform timestep grouping scheme; and
$\textcircled{2}$ introduce a weight-sharing mixed-precision solver to accelerate the evaluation process by reusing the results of PTQ calibration. 
}
\label{fig:framework}
\end{figure*}

In this paper, we propose to simultaneously reduce the time-step length and perform low-precision quantization for the diffusion model. 
Our target is to reduce the troublesome costs of models on both temporal and model dimensions. 
For temporal optimization, we search for the optimal subsequence of the previous full sequence timesteps, accelerating the denosing process by identifying and removing the redundant timesteps. 
For quantization on model optimization, we further adopt mixed-precision quantization for the generative models, \ie, different layers are assigned to different quantization precision according to their quantization sensitivities, while maintaining the overall model compression ratio. 

However, the resulting combinations of these two dimensions grow exponentially, posing significant challenges for efficient and effective optimization and evaluation.
For DDPM model \cite{DDPM}, suggesting each layer has 4 bit-width choices and the expected timestep sequence is to be shortened from 1000 to 20, the resulting total combinations are roughly $ 4^{250} \times10^{42} \approx 3\times10^{192}$, which is already $10^{100}$ times larger than the number of atoms in our universe.

To enhance the efficiency of model optimization, we first design a \emph{weight-sharing precision solver} for fast evaluation of the goodness of a specific layer-wise quantization policy, which eliminates the needless of repetitive calibration during PTQ process. 
For enhancing the temporal efficiency, we conduct comprehensive experiments and observe the timestep exhibits well \emph{non-uniform property}, \ie, timestep contributes unequally to the final generated images. Motivated by this observation, we design a non-uniform time-step grouping scheme for reducing the combinatorial number of possible timestep sub-squence. 
Then, inspired by Neural Architecuture Search (NAS) \cite{tang2023elasticvit,yu2020bignas,li2023diffnas}, we showcase that both timestep reduction and precision selection can be integrated into a unified search space and thus search jointly. 
The overall framework is shown in Fig.~\ref{fig:framework}.

In summary, our contributions are as follows: 
\begin{itemize}
    \item We propose TMPQ-DM, the first method that jointly optimizes the timestep reduction and performs mixed-precision quantization for the diffusion models. 
    \item To better navigate in the large joint space, we devise a \emph{weight-sharing precision solver} for accelerating the evaluation process of mixed-precision quantization, and leverage the \emph{non-uniform property} to group the timesteps for reducing the needless timestep sub-squence. 
    Furthermore, we stuff the search process into a unified search space and perform joint search with a gradient-free evolutionary search algorithm. 
    \item Extensive experiments are conducted on 4 representative datasets with different resolution, including 32$\times$32 CIFAR, 256$\times$256 LSUN-Churches, 256$\times$256 LSUN-Bedrooms, 256$\times$256 ImageNet and 512$\times$512 COCO.
    We achieve more than $10\times$ BitOPs savings on all these tasks while maintaining the same generative performance, showcasing the potential and universality of proposed method. 
\end{itemize}
\section{Related Work}
\subsection{Diffusion Models} 
Different from previous straightly one-step sampling methods such as variational autoencoder, generative adversarial networks, flow models and so on, diffusion models sample gradually from random noise to realistic images.

Given real images $x_0\sim q(x)$, diffusion models~\cite{DDPM} gradually add noise and eventually transform real images to noise as the following equation:
\begin{equation}
    \label{eq:forward_process}
    q(x_t|x_{t-1})=\mathcal{N}(x_t;\sqrt{1-\beta_t} x_{t-1},\beta_t\mathcal{I})
\end{equation}
where $\beta_i\in(0,1)$ is hyper-parameter controlling the diffusion process. Based on markov chain property, Eq. \eqref{eq:forward_process} can then be transformed to:
\begin{equation}
    \label{eq:forward_process_last}
    q(x_t|x_0)=\mathcal{N}(x_t;\sqrt{\overline{\alpha_t}}x_0,(1-\overline{\alpha_t}\mathcal{I})
\end{equation}
where $\alpha_t=1-\beta_t$ and $\overline{\alpha_t}=\prod_{s=1}^t \alpha_s$.

After that, the reverse process is defined to gradually remove noise and eventually recover from noise image to real image under the following equation:
\begin{equation}
    \label{eq:reverse_process}
    q(x_{t-1}|x_t,x_0)=\mathcal{N}(x_{t-1};\widetilde{\mu_t}(x_t,x_0),\widetilde{\beta_t}\mathcal{I})
\end{equation}

In practice, the number of timesteps in diffusion models is usually very huge, up to 1000 steps. Each step needs to forward the model once, which brings a lot of time and computing resource overhead. 
Compared with traditional generative models, diffusion models take more than 1000 times longer to sample the same number of images on the CIFAR-10 dataset \cite{DDIM}, which hinders the application of diffusion models.  

\noindent \textbf{Efficient Sampling Techniques.}
Numerous studies have been devoted to addressing the inefficiencies inherent in the sampling processes of diffusion models from various perspectives, significantly enhancing the speed of diffusion models' sampling operations. 

A class of methods leverage numerical techniques to minimize the number of sampling steps required in diffusion models. Notably, DDIM~\cite{DDIM} circumvents the traditional Markovian constraints imposed on diffusion models by redefining the sampling process's underlying formula, achieving a reduction of up to 90\% in the required sampling timesteps. In addition, PLMS~\cite{PLMS}, DPM-Solver~\cite{dpm-solver} treat diffusion process as solving differential equations (SDE) and give solution from SDE prospective. While these innovations substantially improve the operational efficiency of diffusion models, they rely on pre-defined timestep selection strategies, which may not always be optimal.

A class of methods employ strategic search techniques to optimize sampling efficiency. DDSS~\cite{DDSS} treats the sampling method (\eg, DDIM, DPM-Solver, PLMS) and the selection strategy of different timesteps as a comprehensive search domain.
AutoDiffusion~\cite{Autodiffusion} treats timestep selection and skipping strategies in model layers as search space. While these approaches have succeeded in formulating more efficient sampling strategies, there remains an unexplored potential for enhancing sampling efficiency within the realm of quantization, a gap yet to be addressed by existing research. 

\subsection{Quantization} 
In this paper, we consider the context in post-training quantization (PTQ), as it can achieve higher efficiency than quantization-aware training (QAT) \cite{nagel2020up,hubara2021accurate} and avoid substantial training costs for performing finetuning. 

Quantization aims to map the full-precision inputs $\mathbf{v}$ to low-precision quantized outputs $\mathbf{v}_q$, thus the storage and computation costs can be cut down significantly. 
The quantization function, also known as the quantizer, is used for performing uniform quantization \cite{choi2018pact,esser2020learned} rather than non-uniform quantization \cite{li2019additive,wang2021adaptive}, due to uniform quantization is more hardware-friendly. 
For a bit-width $b$, the uniform quantization function is typically parameterized by a scaling $s$ and a zero-point $z$ for activation quantization is as follows: 
\begin{align}
    \mathbf{v}_q = Q(\mathbf{v};b, s, z) = s \cdot \left( \text{clip} (\nint*{\mathbf{v} \big/ s} + z, 0, 2^{b} - 1 ) - z \right), 
    \label{eq:quantizera}
\end{align}
and the uniform quantizer for weight activation is as follow: 
\begin{small}
\begin{align}
    \mathbf{v}_q = Q(\mathbf{v};b, s, z) = s \cdot \left( \text{clip} (\nint*{\mathbf{v} \big/ s} + z, -2^{b-1}, 2^{b-1} - 1 ) - z \right).  
    \label{eq:quantizerw}
\end{align}
\end{small}
For PTQ, a block-wise reconstruction loss, $\mathop{\arg\min}_{s,z} ||\hat{f}^{(l)}_{b}(x) - f^{(l)}(\mathbf{v}_q)||_2 $, is applied for optimizing the scaling and zero-point, where $f^{(l)}(\cdot)$ indicate the output of full-precision $i-$th block (\eg, the bottleneck block), and $\hat{f}^{(l)}_{b}(x)$ indicate the output of quantized $i-$th block under $b$ bits.

\noindent \textbf{Mixed-Precision Quantization.} 
Mixed-Precision Quantization (MPQ) \cite{tang2024retraining,wang2019haq,guo2020single,cai2020rethinking,wang2021generalizable} is a technique to achieve low-precision quantization, acknowledging the intrinsic diversity in redundancy across various layers within the architectures of deep model. 
Through assigning lower bit-widths to layers exhibiting pronounced redundancy and keeping higher bit-widths to layers having more contributions to final performance, MPQ endeavors to compress model complexity whilst mitigating any notable degradation in performance. 
The challenge resides in assigning the optimal bit-width for individual layers, given the discrete nature of bit-width selection and the exponentially expanding realm of potential policies, encompassing distinct combinations of bit-widths and layers. 

Allocator-based methods \cite{wang2019haq, elthakeb2020releq} adopt reinforcement-learning (RL) to learn the policy assignment agent, and then use this agent to allocate the layer-wise precision. 
Inspired by neural architecure search (NAS), differential-based approaches \cite{cai2020rethinking,yu2020search,tang2023seam,wang2021generalizable} aim to design low-cost weight-sharing methods to learn the probability of bit-width for a given computational resource constraint. 
Furthermore, indicator-based techniques \cite{dong2019hawq,dong2020hawq,chen2021towards,tang2022mixed} use heuristic methods (\eg, Taylor expansion \cite{dong2019hawq,dong2020hawq}, scaling factors of quantizer \cite{tang2022mixed}, \etc) to construct the quantization sensitivity proxy for assisting bit-width assignment. 
However, the above MPQ methods are designed to discriminative models (\eg, CNNs for classification tasks).

\subsection{Neural Architecture Search} 
Neural Architecture Search  (NAS)~\cite{howard2019searching,tan2019efficientnet,zoph2016neural} automates the process of deep models design, which typically consists of several operators (\eg, kernel size, channels, \etc) in a well-defined search space. 

Earliest NAS research resorts to reinforcement learning (RL), which is hampered by the inherent exploration-exploitation dilemma, restricting their exploration to diminutive datasets and consequently yielding suboptimal generalization performance of the discovered architectures \cite{guo2020single,zoph2016neural}. 
To generalize NAS to real-world scenarios, recent works \cite{guo2020single,tang2023elasticvit,yu2020bignas} have focus on the utilization of super-networks. These super-networks leverage weight-sharing mechanisms, wherein subsets of weights within the search space are shared among architectures.
This significantly shortens the evaluation time of the goodness for a given architecture \cite{liu2018darts,pham2018efficient}. 
In this paper, we leverage this paradigm to the optimization of quantization precision for individual layers and the determination of optimal subsequences of timesteps, with the overarching goal of concurrently optimizing temporal and model efficiency. 
\section{Method}
In this section, we introduce several techniques in Sec \ref{sec:foundation} to provide a foundation for conducting efficient optimization of diffusion models. 
Specifically, we first introduce the mixed-precision solver for fast evaluation of a specific quantization policy to avoid repetitive calibration of PTQ. 
Then, we design a non-uniform timestep grouping scheme to achieve search space reduction for fast optimization, which is motivated by the observation that timesteps contribute differently to the final generation quality. 

\subsection{Foundation of Efficient Optimization} 
\noindent \textbf{Mixed-Precision Solver. } 
\label{sec:foundation}
To identify the optimal mixed-precision quantization settings for each layer, we should evaluate its performance under different resource constraints (\eg, BitOPs, latency, \etc.). 
However, it is non-trivial to efficiently obtain the direct performance because the calibration process for post-training quantization is costly. 
Moreover, the intricate architectural composition of diffusion models, characterized by the presence of encoder and decoder blocks, leads to a higher layer count relative to conventional discriminative models, thereby intensifying the computational demands inherent in this endeavor. 

\begin{figure}
    \centering
    \includegraphics[width=0.45\textwidth]{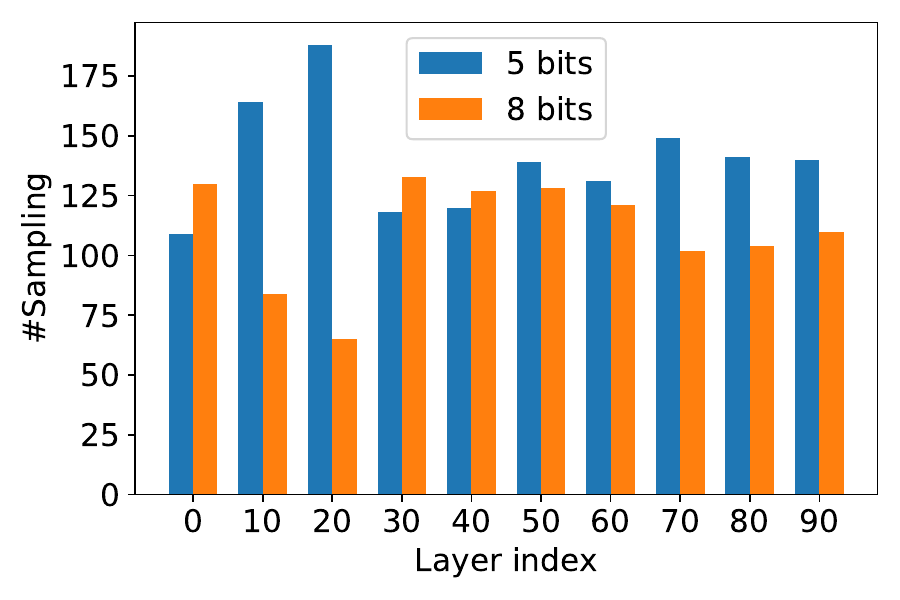}
    \caption{Number of shared precision under 500 randomly sampled mixed-precision quantization policies under same BitOPs constraint. 
    }
  \label{fig:sampling_prob}
  \vspace{-0.6cm}
\end{figure}

Luckily, we have observed different mixed-precision settings could share some \emph{same precision}. 
Specifically, for $K$ mixed-precision quantization settings $\left \{ \{(w_l^0, a_l^0)\}_{l=0}^{L-1}, ..., \{(w_l^K, a_l^K)\}_{l=0}^{L-1} \right\}$, we can always find there has overlapped precision for same layer $i$ of across different MPQ settings, \eg, $w_i^A = w_i^B = w_i^C$ ..., where $A \neq B \neq C  $. 
As shown in Fig. ~\ref{fig:sampling_prob}, we randomly sample 500 non-overlapping mixed-precision settings and count the shared precision of 5bits and 8bits across layers. 
One can observe that there are several repetitive precision selection patterns of layers, suggesting that it is possible to avoid repeated calibration by reusing the results of single calibration process. 
Hence, we introduce the mixed-precision solver, a pre-build super-network that share the calibration results and thus only needs one-time calibration for serving as a accurate performance indicator. 

To build the precision solver, we first convert the original quantizer in Eq. \eqref{eq:quantizera} and \eqref{eq:quantizerw} to support multi-precision. 
That's to say, each quantizer will have $N$ quantization parameters of scaling and zero-points to match $N$ possible bit-widths, resulting in a scaling set $S=\{s_0,...,s_{N-1}\}$ and zero-point set $Z=\{z_0,...,z_{N-1}\}$, where each $\{z_i, s_i\}$ corresponding to a specific bit-width for this quantizer. 
At each optimization iteration, for $j-th$ block ${F}^{(j)}$ with $k$ layers, \ie, ${F}^{(j)}=f^{\{(j), (k-1)\}} \circ ... \circ f^{\{(j), (0)\}} (x^{(j-1)})$ , where $x^{(j-1)}$ is the outputs of last block, we sample one bit-width $b$ from the bit-width candidates and switch the quantization parameters of each quantizer to the corresponding bit-width for the forward, and then update the quantization parameters in the backward process for the layers in this block on a small-scale calibration dataset, with the following optimization objective: 
\begin{equation} 
    \label{eq:mixed-quantization-solver}
    \mathop{\arg\min}_{\small {\{S_l\}}_{l=0}^{k-1}, {\{Z_l\}}_{l=0}^{k-1}} || \hat{F}^{(j)}_{b} (x^{(j-1)}) - F^{(j)}(x^{(j-1)}) ||_2. 
\end{equation} 
During this process, the parameters of the backbone model is fixed, we only update the quantization parameters for the  quantizers, as in conventional post-training quantization. 
After all bit-widths are sampled and the corresponding quantization parameters in $j-$th block are updated, we switch to next $(j+1)-$th block and  repeat the above procedure. 
The forward and backward processes are shown in Fig. \ref{fig:framework} (a). 

\noindent \textbf{Non-uniform Timestep Grouping.}
\begin{figure}[tb]
  \centering
  \begin{subfigure}{0.45\linewidth}
    \includegraphics[width=\textwidth]{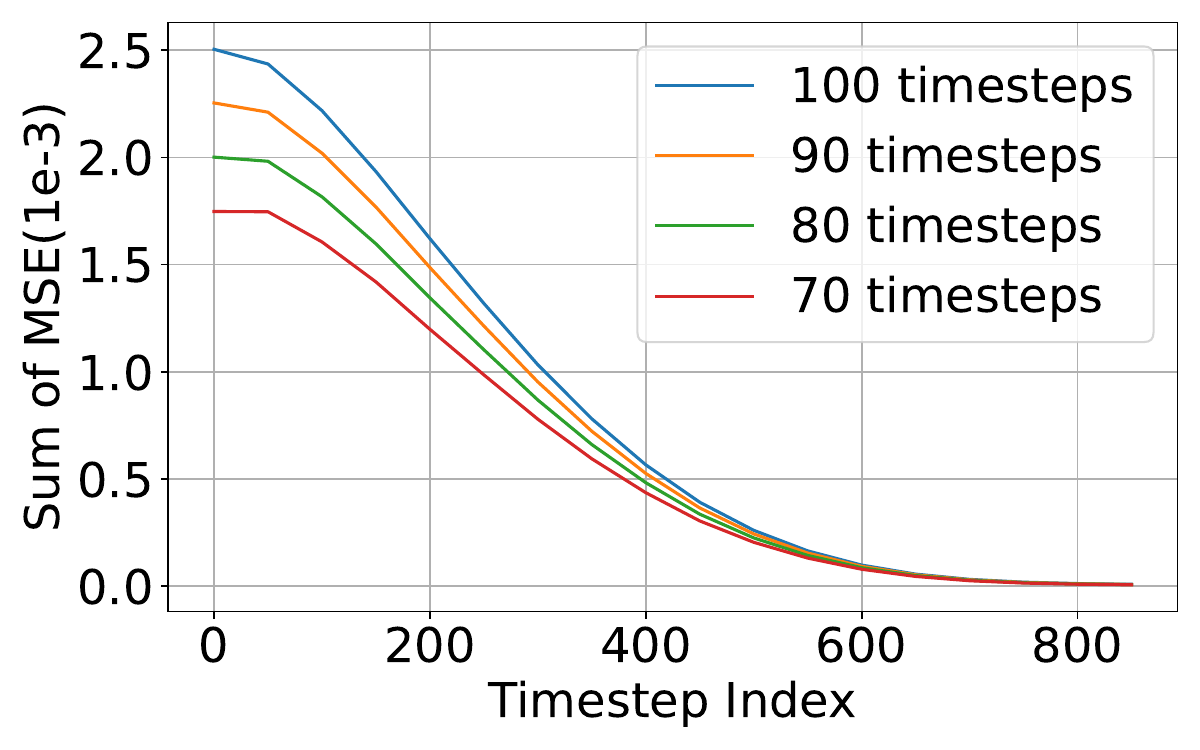}
    \caption{Quantized}
    \label{fig:quantized_mse}
  \end{subfigure}
  \hfill 
  \begin{subfigure}{0.45\linewidth}
    \includegraphics[width=\textwidth]{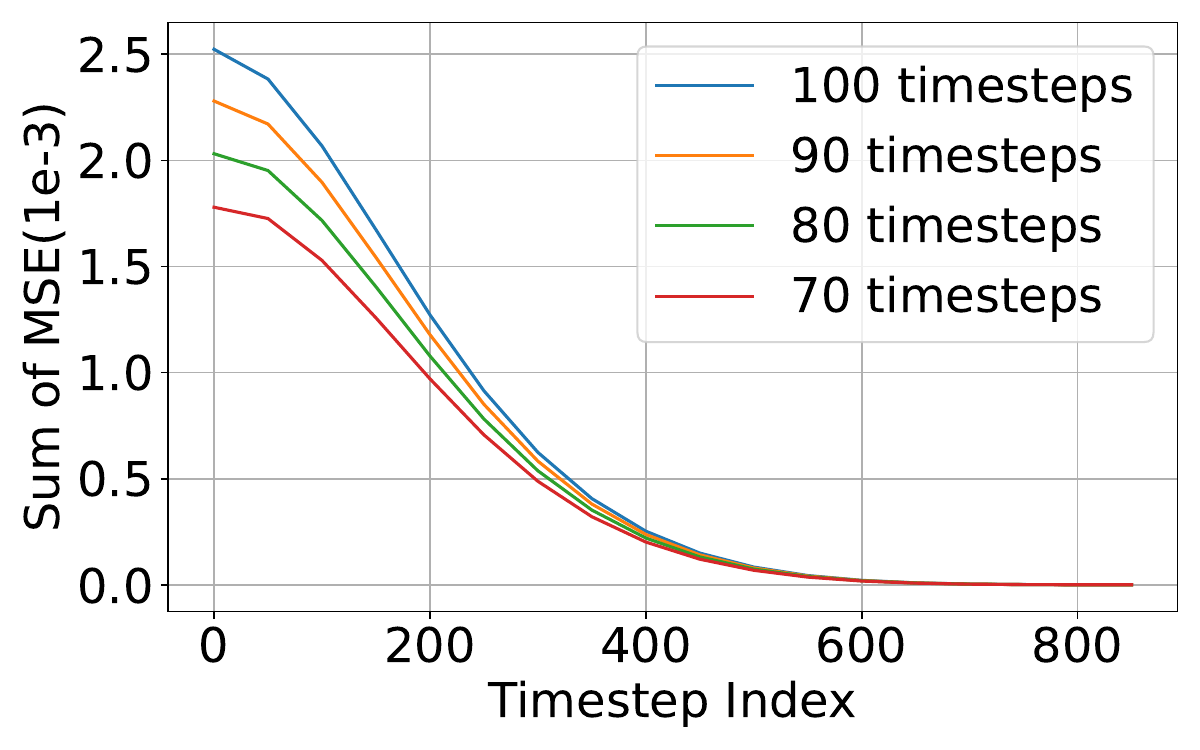}
    \caption{Full Precision}
    \label{fig:fp_mse}
  \end{subfigure}
  
  \caption{MSE for timesteps. The temporal feature difference of quantized model is slightly large than full-precision model, further showcasing it needs a non-uniform grouping scheme. }
  \label{fig:time_step_mse}
\end{figure}
On the other hand, the total number of timestep selection is actually a combinatorial number. For example, if we want to pick 20 timestep out of total 1000 timesteps, the combinatorial number becomes $\binom{1000}{20} = 10^{42}$, which also takes efficiency problem. 

A feasible solution is to group the timesteps with similar denoising behavior, so that we can reduce the daunting timestep scale. 
However, a straightforward scheme--uniform grouping--might be sub-optimal. 
As shown in Fig.~\ref{fig:time_step_mse}, we calculate the temporal feature difference $D_{T,t}$ for a given time window $T \in \{70,80,90,100\}$ and current timestep $t$ by 
\begin{equation}
    D_{T,t} = \sum \text{MSE}\left( x_t,x_{t-T} \right) = \sum_{i=1}^T \text{MSE} (x_{t+i-1},x_{t+i}). 
\end{equation}
Our empirical observation in Fig.~\ref{fig:time_step_mse} reveals different divergences in the behaviors exhibited by timesteps. Specifically, smaller timesteps manifest a heightened disparity, attributable to their closer proximity to real images, while larger timesteps evince a comparatively diminished divergence. 
This observation motivate us for the formulation of the subsequent non-uniform grouping scheme: 
\begin{equation}
g(t)=\begin{cases}
h,\quad (\sqrt{0.8T}\frac{h-1}{H-1})^2\le t < (\sqrt{0.8T}\frac{h}{H-1})^2 \\
H,\quad t\ge 0.8T \\
\end{cases},
\end{equation}
where $T$ denotes the total number of timesteps, $H$ represents the number of timesteps selected to sample, $t$ represents the index of timestep. The function 
$g(t)$ maps each timestep $t$ to a group index while $h$ serves as the aforementioned group index where $t<0.8T$.   

\subsection{Unified Configuration Search} 
Inspired by NAS that encodes the components of interests into a search space and search accordingly, we describe our search space and the resource-constrained search as follows: 

\noindent \textbf{Search Space.} 
For precision optimization, we consider the weight and activation of each layer have independent bit-width candidates. 
For each layer, $\mathbf{B}_{weight}=\{b_i\}_{i=0}^{M-1}$ and $\mathbf{B}_{act}=\{b_j\}_{j=0}^{N-1}$ represent the weight and activation candidates, respectively. 
Therefore, each linear/convolution layer has $M*N$ bit-width choices. 
For timestep optimization, we stuff the timesteps into $K$ disjoint groups, where $K$ is the expected number of timesteps. Then, we search a most representative timestep for each group. 

\noindent \textbf{Resource-constrained Search.} 
We perform evolutionary search over the search space $\mathcal{A}$ under a given resource constraint. 
In this paper, we calculate the integer operations for each timestep as the constraint, represented by Bitwise Operations (BitOPs) \footnote{We also quantize the inputs of matrix multiplication in self-attention layers, which are two activations (feature maps), the resulting BitOPs is calculated as ${\rm BitOPs} = {\rm\#MACs} \times b_a \times b_a$.}: 
\begin{equation}
    {\rm BitOPs} = {\rm\#MACs} \times b_a \times b_w , 
\end{equation}
where ${\rm\#MACs}$ is the multiply-accumulate operations. 
Those configurations exceed constraint will be discarded before evaluation. 

For search, we first initialize with $C$ randomly sampled configurations and generated a small number of samples to calculate FID to evaluate configurations as follows: 
\begin{equation}
    \text{FID} (R, G) = ||\mu_R-\mu_G||_2 ^ 2 + Tr(\Sigma_R + \Sigma_G - 2(\Sigma_R\Sigma_G)^\frac{1}{2}), 
\end{equation}
where $R$, $G$ represent real images and generated images, $\mu$ and $\Sigma$ represent the mean and covariance of the feature distribution in the last pooling layer of Inception V3 model for real and generated images. $Tr(\cdot)$ represents the trace of the matrix. 
During each epoch of search, we get new configurations through crossover, mutation and random generation operations and evaluate corresponding FID to update a Top-k configuration set. 
After all configurations in this epoch get evaluated, we choose Top-k configurations with lowest FID, which means better performance, as parent configurations for next epoch to choose from. 
To get crossover configurations, we randomly select two parent configurations and choose timesteps and bitwidth allocations evenly from them. To get mutation configurations, we randomly select one parent strategy and change timesteps and bitwidth allocations with $p$ probability. 

Moreover, we conduct offline sampling of the random mixed-precision settings to alleviate the heavy time costs of random sampling within expected resource constraints. 
Specifically, we sample different bitwidth allocation configurations using CPU before the evolutionary search that meet our BitOPs constraints, which can be done on multi-core CPUs with multi-processes parallel processing under different random seeds for acceleration. 
\section{Experiment}

\subsection{Experiment Setting}

\noindent \textbf{Models and Datasets.} 
\label{sec:4.1.1}
To showcase the adaptability of our method, we evaluate TMPQ-DM on the original DDPM~\cite{DDPM} using CIFAR-10 dataset, Latent Diffusion model~\cite{LatentDiffusion} on LSUN-Bedrooms, LSUN-Churches, and ImageNet datasets. 
We also experiment for text-guided image synthesis on the COCO dataset. 
We conduct experiments with frequently-used fast sampling techniques. 
Specifically, DDIM is utilized for the CIFAR-10, LSUN-Bedrooms, LSUN-Churches, and ImageNet datasets, whereas DPM-Solver is applied to the COCO dataset experiments.

\noindent \textbf{Quantization Settings.}
We use Q-diffusion \cite{li2023q} as base quantization method. We compare our results with ADPDM \cite{ADPDM} which searches for 8 groups of quantization parameters and assigns one of these groups to each timestep. 
We quantize all linear and convolution layers as in previous work~\cite{he2024ptqd,shang2023post}. 
Moreover, we quantize all input matrices involved in the matrix multiplications within the attention blocks, specifically including queries, keys, and values.

\noindent \textbf{Hyper-parameters.} 
\label{sec:4.1.2}
We employ evolutionary algorithm to identify optimal strategies. For evolutionary algorithm, the population size $P = 50$, the mutation number $m = 25$ and the cross number $c = 10$ per epoch, with the mutation probability set at 0.25. The search spans 20 epochs, during which we collect 2000 samples for policy prediction on the CIFAR dataset. For the LSUN-Bedrooms, LSUN-Churches, ImageNet, and COCO datasets, the search is conducted over 10 epochs with 1000 samples collected. 
We set quantization bit-width candidates to $\mathbf{B}_{weight} = \mathbf{B}_{act} = \{5, 6, 7, 8\}$.

\begin{table*}[t]
  \caption{FID and IS for DDPM on CIFAR-10 ($32\times32$) in different settings, varying the number of timesteps, ``TS'' denotes ``Timestep Search'', ``MP'' denotes ``Mixed Precision'', ``GS'' denotes ``Groupwise Splition''.}
  \label{tab:cifar}
  \centering
  \begin{tabular}{@{}cccccc@{}}
    \toprule
    Steps & Scheme &Bit-width& \makecell[c]{Overall \\ BitOPs (T)}  & FID ($\downarrow$) &IS ($\uparrow$)\\ 
    \midrule
    100 & Q-diffusion&$6/6$     & $44.3$       & 33.98 & 8.26\\
    100 & ADPDM    & $6/6$&   $44.3$&6.57 &9.06\\\cline{1-6}
    5 & Q-diffusion  &$6/6$    &     $2.21$          & 68.93 & 5.67\\
    5 & +TS &$6/6$&$2.21$& 46.04$\pm$0.30 & 7.17$\pm$0.11 \\
    5 & ++MP &$6_{MP}/6_{MP}$ & \textbf{$2.17$} & \textbf{13.08}$\pm$0.32 & 8.35$\pm$0.11  \\ 
    5 & +++GS &$6_{MP}/6_{MP}$ & $2.12$ & 15.48 $\pm$0.64 & \textbf{8.46} $\pm$0.12  \\ \cline{1-6}

    10 & Q-diffusion&$6/6$& $4.43$ & 37.80 & 7.42 \\
    10 & +TS &$6/6$&$4.43$ & 30.38 $\pm$0.22 & 8.07$\pm$0.1  \\
    10 & ++MP &$6_{MP}/6_{MP}$& \textbf{$4.29$} & 9.57 $\pm$0.28 & 8.8$\pm$0.1  \\ 
    10 & +++GS &$6_{MP}/6_{MP}$& $4.35$ & \textbf{8.76} $\pm$0.25 & \textbf{9.08}$\pm$0.11  \\ \cline{1-6}

    15 & Q-diffusion&$6/6$&$6.64$& 34.12 & 7.82 \\
    15 & +TS &$6/6$&$6.64$ & 27.70 $\pm$0.12 & 8.47$\pm$0.13  \\
    15 & ++MP &$6_{MP}/6_{MP}$& \textbf{$6.38$} & 6.94 $\pm$0.41 & 8.93$\pm$0.09  \\ 
    15 & +++GS &$6_{MP}/6_{MP}$& $6.54$ & \textbf{5.28} $\pm$0.25 & \textbf{9.19} $\pm$0.12 \\ 
  \bottomrule
  \end{tabular}
  
\end{table*}

\noindent \textbf{Evaluation Metrics.} 
\label{sec:4.1.3}
We evaluate FID across all experiments, which measures the distributional disparity between generated and real images. 
For CIFAR-10 and ImageNet datasets, IS is additionally applied to gauge the quality of the images generated. To evaluate inference efficiency, we calculate the Bit Operations (BitOPs) which measures the total number of bitwise operations required. 
We denote the complexity of the model during once inference (\ie, single timestep) as BitOPs and use the term \underline{Overall BitOPs} (\ie, the cumulative BitOPs across all timesteps) to represent the total complexity involved in generating a full image : 
\begin{equation} 
\scriptsize 
    \text{Overall BitOPs} = \underbrace{\text{BitOPs}}_{\mbox{cost of single timestep}} \times  \underbrace{\text{Number of timesteps}}_{\mbox{length of timestep sequence}} . 
\end{equation}
For Latent Diffusion, we focus solely on the diffusion processes in the latent space, excluding the encoding and decoding steps from our calculations to concentrate on diffusion-specific issues.

\begin{figure*}[htbp]
    \centering
    \begin{subfigure}{0.475\textwidth}
    \centering
    \includegraphics[width=\textwidth]{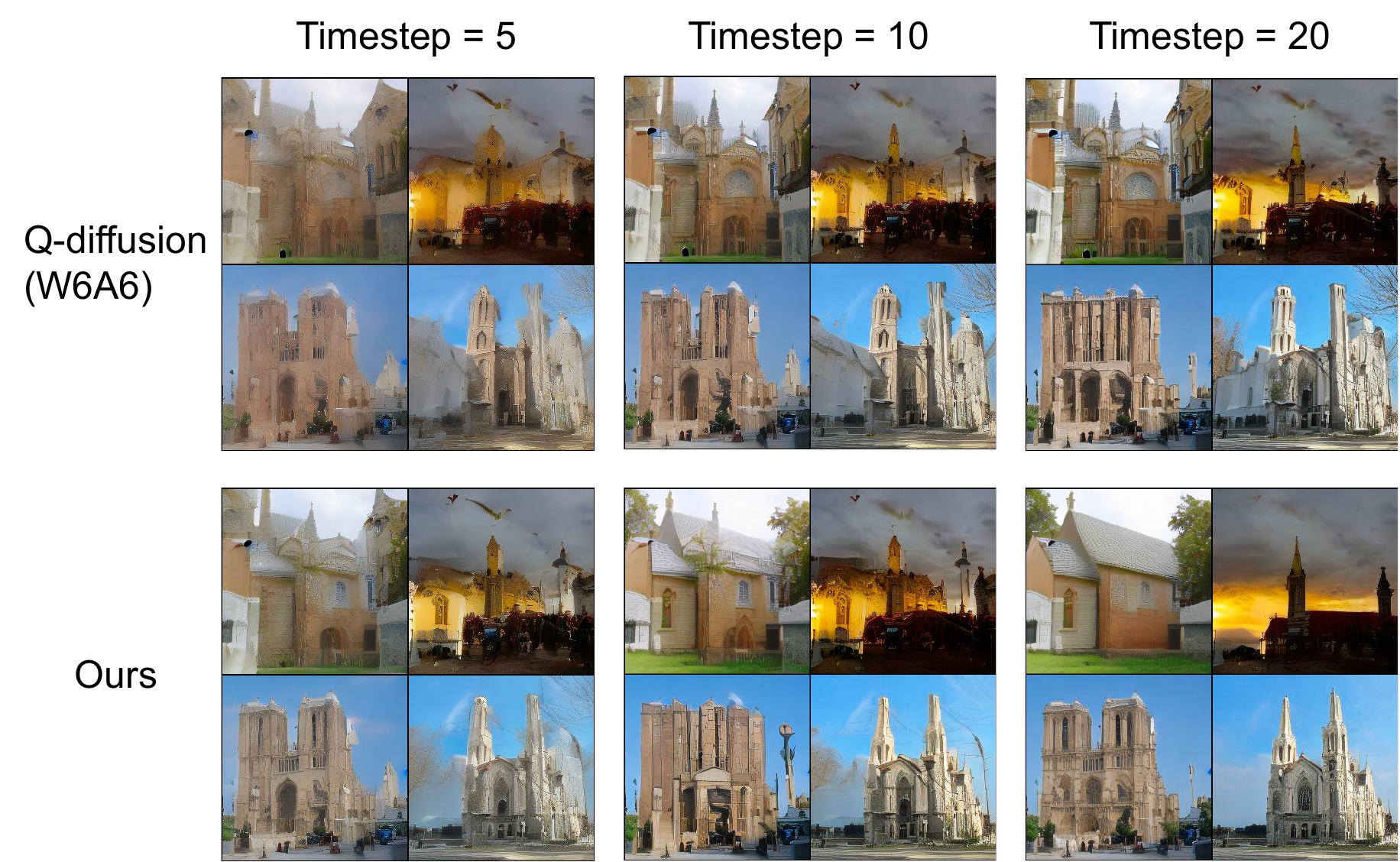}
        \caption{Visualization of LSUN-Churches}
        \label{fig:subfig1}
    \end{subfigure}
    \hfill
    \begin{subfigure}{0.475\textwidth}
    \centering
    \includegraphics[width=\textwidth]{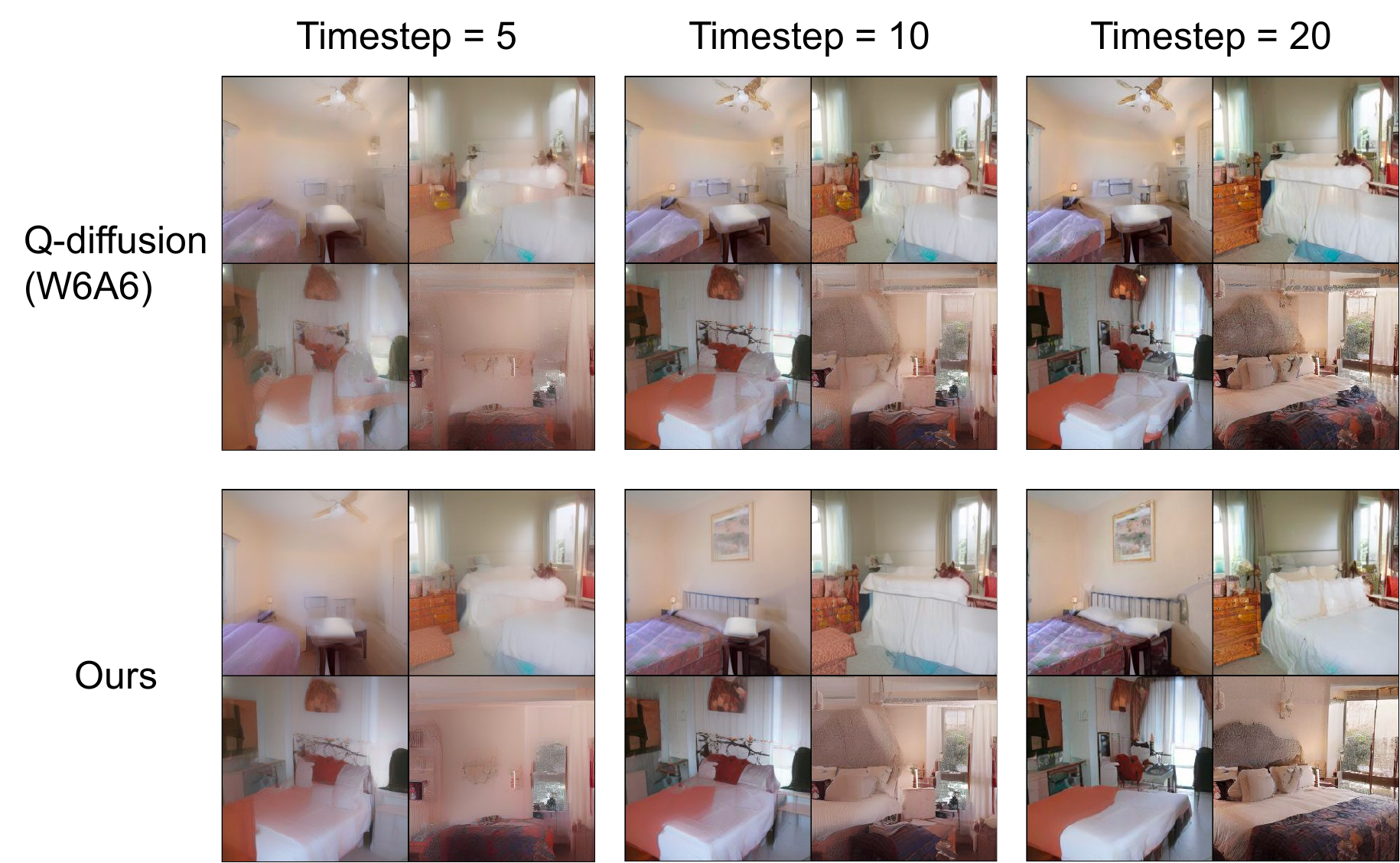}
        \caption{Visualization of LSUN-Bedrooms}
        \label{fig:subfig2}
    \end{subfigure}
    \caption{Unconditional Image Generation with or without our method under different timesteps. Compared to direct application of Q-diffusion, our approach can generate more realistic images.}
    \label{fig:LSUN}
\end{figure*}

\subsection{Main Results}
\label{sec:4.2}
\noindent \textbf{Low-resolution Image Generation.} 
We initially conduct experiments on the CIFAR-10 dataset using the original DDIM-accelerated DDPM with 100 steps~\cite{DDIM} as our baseline. We assess both the Frechet Inception Distance (FID) and the Inception Score (IS) for CIFAR-10. We evaluate the top ten strategies searched and calculated their mean and standard deviation.

As shown in Tab.~\ref{tab:cifar}, experimenting with timestep optimization improves performance to a degree, enhancing the FID from 68.93 to 46.04 and the IS from 5.67 to 7.17 at 5 steps. Furthermore, employing mixed precision searching significantly refines sampling quality, reducing the FID from 46.08 to 13.08 and increasing the IS from 7.17 to 8.35 at the same step count. While at 5 steps non-uniform grouping method does not enhance FID, it does improve IS performance and boosts sampling quality where at 10 or 15 steps. Specifically, our method enhances FID from 6.94 to 5.28 and IS from 9.08 to 9.19 at 15 steps. Notably, with 15 steps, our approach achieves better FID and IS metrics compared to ADPDM with 100 steps, thus saving 85\% in model invocations.

\begin{table}[t]
\fontsize{8}{10}\selectfont
  \caption{FID scores for Latent Diffusion on LSUN-Churches $256\times256$ in different settings under BitOPs constraint, varying the number of timesteps. ``Ours ($ {\small DPM}$)'' represents experimentation with DPM-Solver.}
  \label{tab:church1}
  \centering
  \begin{tabular}{@{}ccccc@{}}
    \toprule
    Steps & Scheme &Bit-width &  \makecell[c]{Overall \\ BitOPs (T)} & FID ($\downarrow$)\\ 
    \midrule\addlinespace
    400 & Q-diffusion &$6/6$&575& 7.83 \\
    100 & ADPDM &$6/6$&144&6.9\\\cline{1-5}
    
    5 & Q-diffusion &$6/6$&7.19& 41.13 \\
    
    5 & Ours &$6_{MP}/6_{MP}$&\textbf{7.13}& \textbf{21.01} \\ \cline{1-5}

    10 & Q-diffusion &$6/6$&14.4& 16.85 \\
    
    10 & Ours &$6_{MP}/6_{MP}$&\textbf{13.8}& \textbf{9.86}\\ \cline{1-5}

    20 & Q-diffusion &$6/6$&28.8& 10.60 \\
    
    20 & Ours &$6_{MP}/6_{MP}$&\textbf{28.6}& 6.22 \\ 
    
    20 & Ours &$7_{MP}/7_{MP}$&36.4& 5.38 \\
    
    20 & Ours ($ {\small DPM}$) &$6_{MP}/6_{MP}$&28.7&\textbf{4.88} \\
  \bottomrule
  \end{tabular}
\end{table}

\noindent \textbf{Unconditional High-resolution Image Generation for Latent Diffusion.} 
To demonstrate our method's effectiveness in generating high-resolution images, we conduct experiments on the $256\times256$ LSUN-Bedrooms and LSUN-Churches datasets. The original DDIM-accelerated Latent Diffusion, set at 400 steps for LSUN-Churches and 200 steps for LSUN-Bedrooms, served as our baseline. For these experiments, we set the $\eta$ parameter of DDIM to 0.

As shown in Tab.~\ref{tab:bedroom1} and Tab.~\ref{tab:church1}, our method substantially enhances image generation quality and consistently surpasses the baseline single-precision quantization method across both datasets and all timestep settings. Notably, under BitOPs constraint of W6A6, our method at 10 timesteps outperforms the baseline at 20 timesteps, significantly reducing inference costs. Furthermore, our approach at 10 timesteps for LSUN-Bedrooms and 20 timesteps for LSUN-Churches achieves results comparable to ADPDM at 100 timesteps, marking a substantial improvement. 
Visualizations are provided in Fig.~\ref{fig:LSUN}. It is evident that images generated with our method exhibit greater fidelity compared to those produced without it. Additionally, our method demonstrates its effectiveness by generating images at 10 timesteps that are comparable in quality to those generated without our method at 20 timesteps.

\begin{table}[t]
\fontsize{8}{10}\selectfont
  \caption{FID scores for Latent Diffusion on LSUN-Bedrooms $256\times256$ in different settings under BitOPs constraint, varying the number of timesteps. ``Ours ($ {\small DPM}$)'' represents experimentation with DPM-Solver.}
  \label{tab:bedroom1}
  \centering
  \begin{tabular}{@{}ccccc@{}}
    \toprule
    Steps & Scheme & Bit-width & \makecell[c]{Overall \\ BitOPs (T)} & FID ($\downarrow$)\\ 
    \midrule
    200 & Q-diffusion &$6/6$&1426& 8.71 \\
    100 & ADPDM &$6/6$&713&9.88\\\cline{1-5}
    
    5 & Q-diffusion & $6/6$ &35.7 &66.26 \\
    5 & Ours &$6_{MP}/6_{MP}$& \textbf{35.0}&\textbf{22.33} \\ \cline{1-5}

    10 & Q-diffusion &$6/6$&$71.3$& 18.89 \\
    10 & Ours &$6_{MP}/6_{MP}$&\textbf{71.2}& \textbf{9.43} \\ \cline{1-5}

    20 & Q-diffusion &$6/6$&$143$& 12.03 \\
    20 & Ours &$6_{MP}/6_{MP}$&137&7.89 \\ 
    20 & Ours &$7_{MP}/7_{MP}$&183&5.44 \\
    20 & Ours ($ {\small DPM}$) & $6_{MP}/6_{MP}$&\textbf{137}&\textbf{5.12}\\
  \bottomrule
  \end{tabular}
\end{table}

\begin{figure*}[t]
  \centering
  \begin{subfigure}{0.3\linewidth}
    \includegraphics[width=\textwidth]{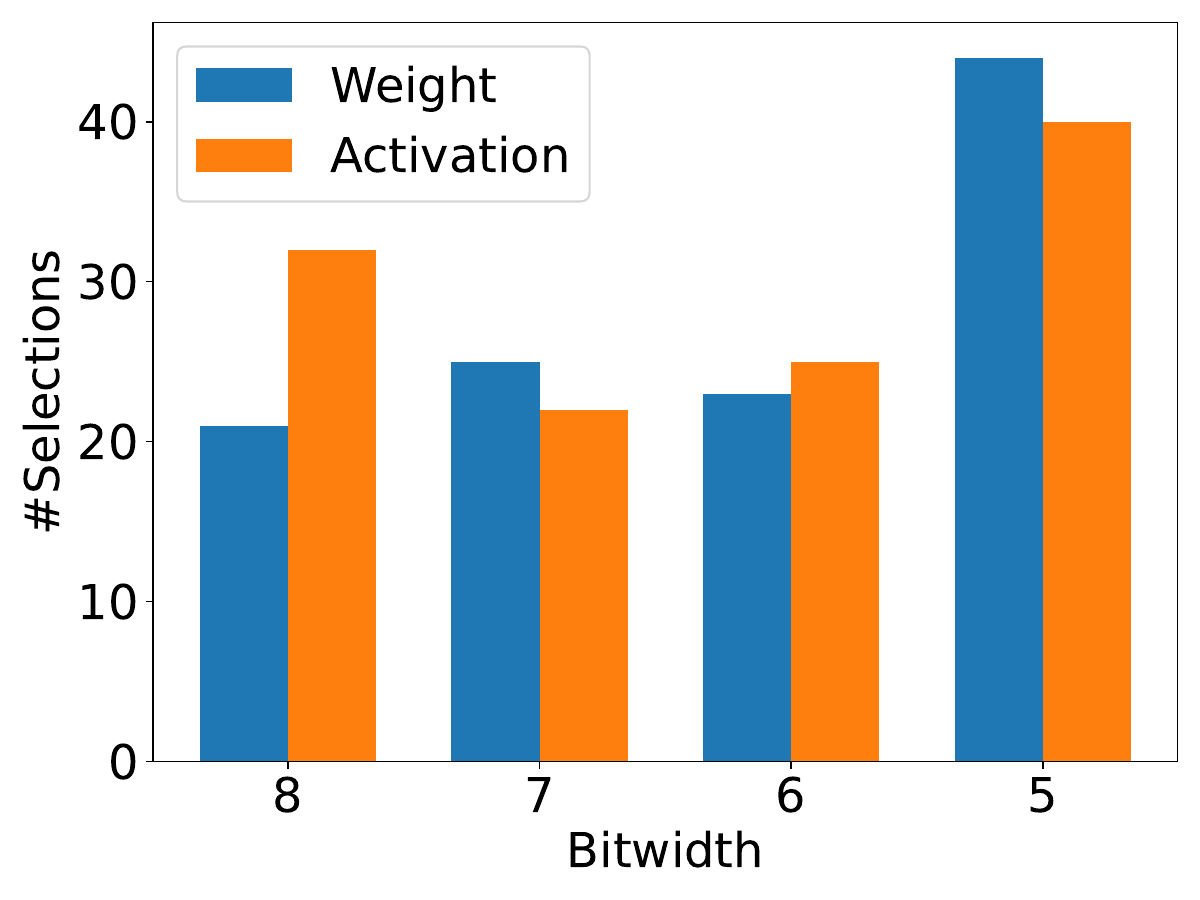}
    \caption{Bit-width Distribution}
    \label{fig:bit_width_distribution}
  \end{subfigure}
  \hfill 
  \hspace{-4cm}
  \begin{subfigure}{0.3\linewidth}
    \includegraphics[width=\textwidth]{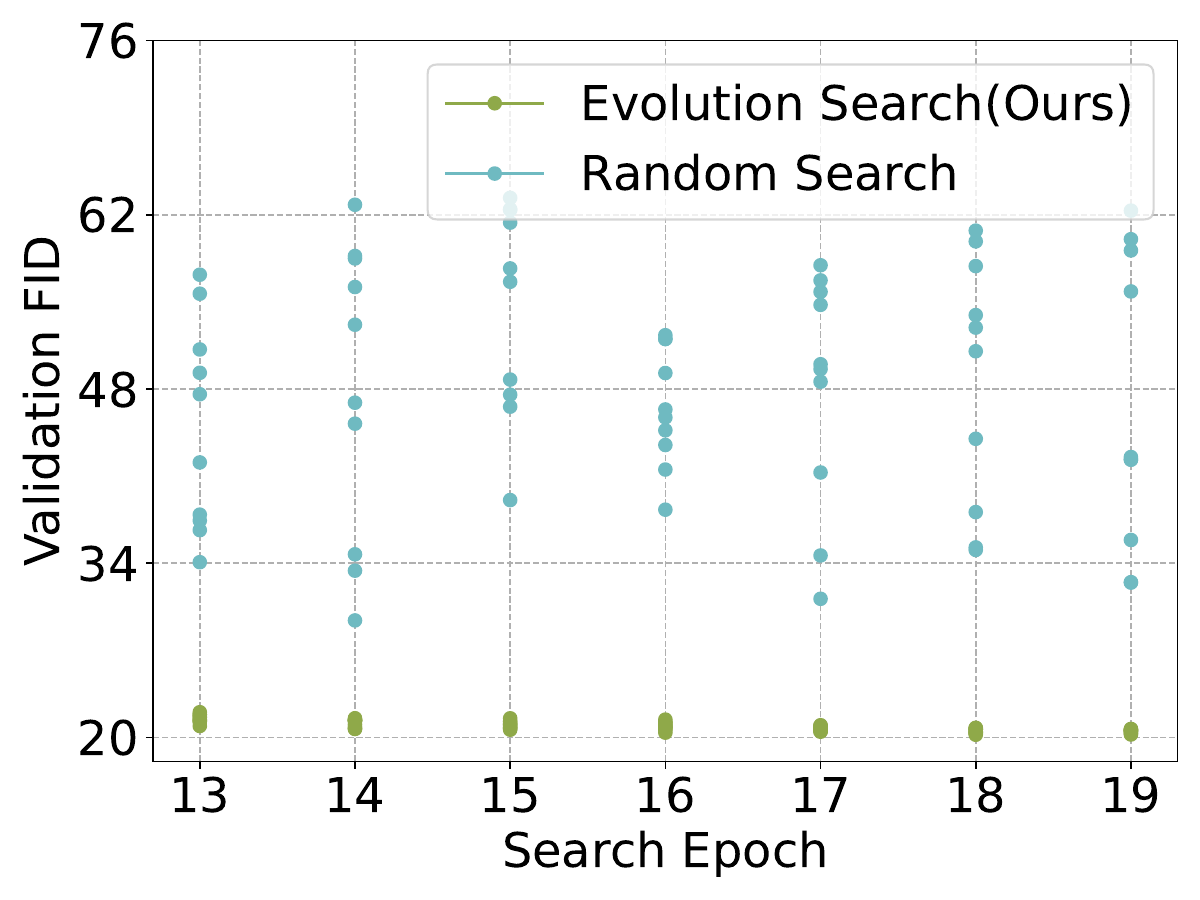}
    \caption{Search Process}
    \label{fig:search_process}
  \end{subfigure}
  \hfill
  \hspace{-4cm}
  \begin{subfigure}{0.3\linewidth}
    \includegraphics[width=\textwidth]{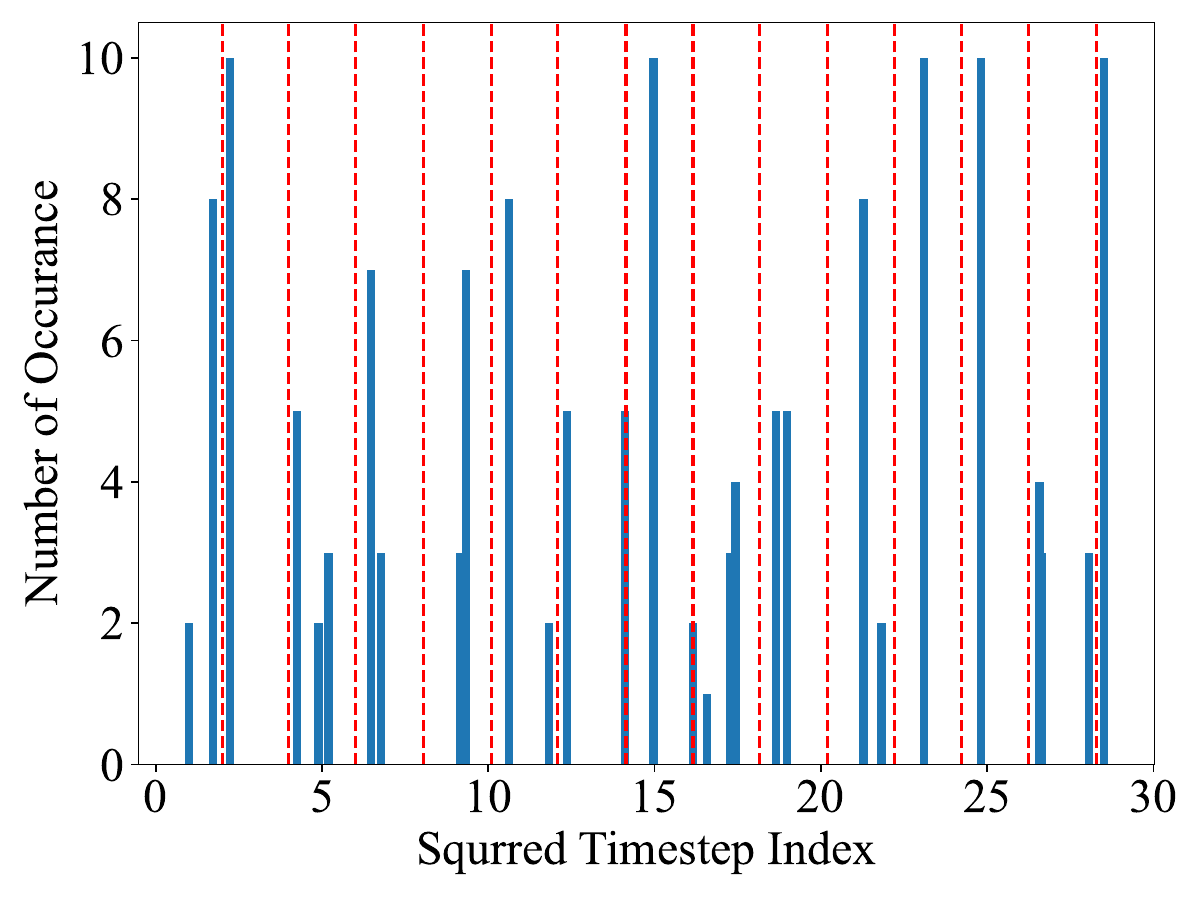}
    \caption{Timestep}
    \label{fig:ts_occ}
  \end{subfigure}
  
  \caption{Activations need higher bit-width than weights (a). Evolutionary search generates better strategies than random search (b). The selection of timestep is mostly positioned in the middle of the grouping (c).}
  \label{fig:exp_comparison} 
\end{figure*}

\begin{table}[t]
\fontsize{8}{10}\selectfont
  \caption{FID, sFID and IS for Latent Diffusion on ImageNet $256\times256$ in different settings under BitOPs of W6A6, varying the number of timesteps.}
  \label{tab:image1}
  \centering
  \begin{tabular}{@{}cccccc@{}}
    \toprule
    Steps & Scheme  & \makecell[c]{Overall \\ BitOPs (T)} & FID ($\downarrow$)&sFID ($\downarrow$)& IS ($\uparrow$)\\ 
    \midrule
    200 & Q-diffusion &1488& 9.44&13.42&316.94 \\
    100 & ADPDM &  744 &11.584&-&178.64\\\cline{1-6}
    6 & Q-diffusion &44.6& 28.27&25.50&135.29 \\
    6 & Ours &\textbf{43.6}& \textbf{12.14}&\textbf{14.40}&\textbf{229.17} \\ \cline{1-6}

    8 & Q-diffusion &59.5& 14.73&21.47&207.72 \\
    8 & Ours & \textbf{59.2}&\textbf{11.76}&\textbf{15.54}&\textbf{228.29} \\ \cline{1-6}

    10 & Q-diffusion &74.4& 10.96&19.58&\textbf{244.38} \\
    10 & Ours &\textbf{74.3}& \textbf{10.41}&\textbf{13.02}&242.17 \\ 
  \bottomrule
  \end{tabular}
\end{table}

\noindent \textbf{Conditional High-resolution Image Generation for Latent Diffusion.} 
To assess the performance of our method on conditional generation tasks, we conducted experiments using the LDM-4 model on the $256\times256$ ImageNet dataset. We utilized DDIM as the fast sampling method, setting $\eta$ to 0 and the guidance scale to 3.

\begin{figure}[t]
    \centering
    \includegraphics[width=0.95\linewidth]{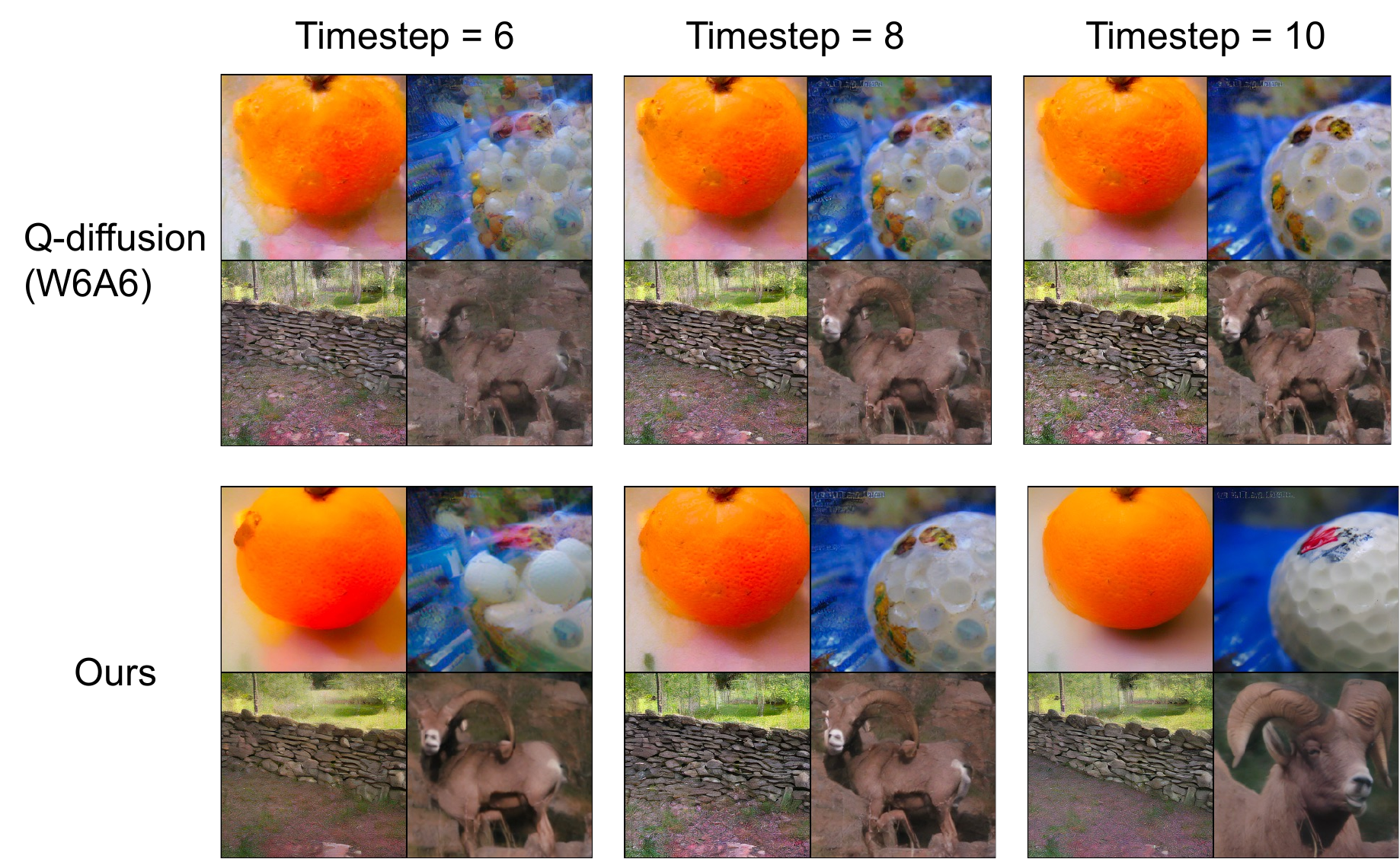}
    \caption{Conditional image generation of Q-diffusion and proposed methods with different timesteps. Our method can generate more realistic images.}
    \label{fig:imagenet}
\end{figure}

Experiments were performed at timesteps of 6, 8, and 10, with evaluations based on FID and IS metrics. 
As demonstrated in Tab.~\ref{tab:image1}, our method consistently enhances performance. Where steps = 6 or 8, our method improves FID, sFID and IS consistently compared to Q-diffusion at equivalent timesteps. Specifically, our approach significantly improves FID and IS from 28.27 and 135.29 to 12.14 and 229.17 respectively at 6 steps.
Visual results, shown in Fig.~\ref{fig:imagenet}, demonstrate that our method generates images with higher fidelity compared to the baseline Q-diffusion at equivalent timesteps.

\noindent \textbf{Text-guided Image Generation for Stable Diffusion.} 
For the text-guided image generation task, we apply our method using the Stable Diffusion model, which was pretrained on the LAION-5B dataset. We generate images corresponding to text prompts to calibrate and refine our search process. Consistent with prior studies, we set the guidance scale to 7.5~\cite{li2023q,LatentDiffusion}. We choose DPM-Solver as the fast sampler, implementing its second-order version as used in Stable Diffusion.

\begin{figure}[t]
    \centering
    \includegraphics[width=0.7\linewidth]{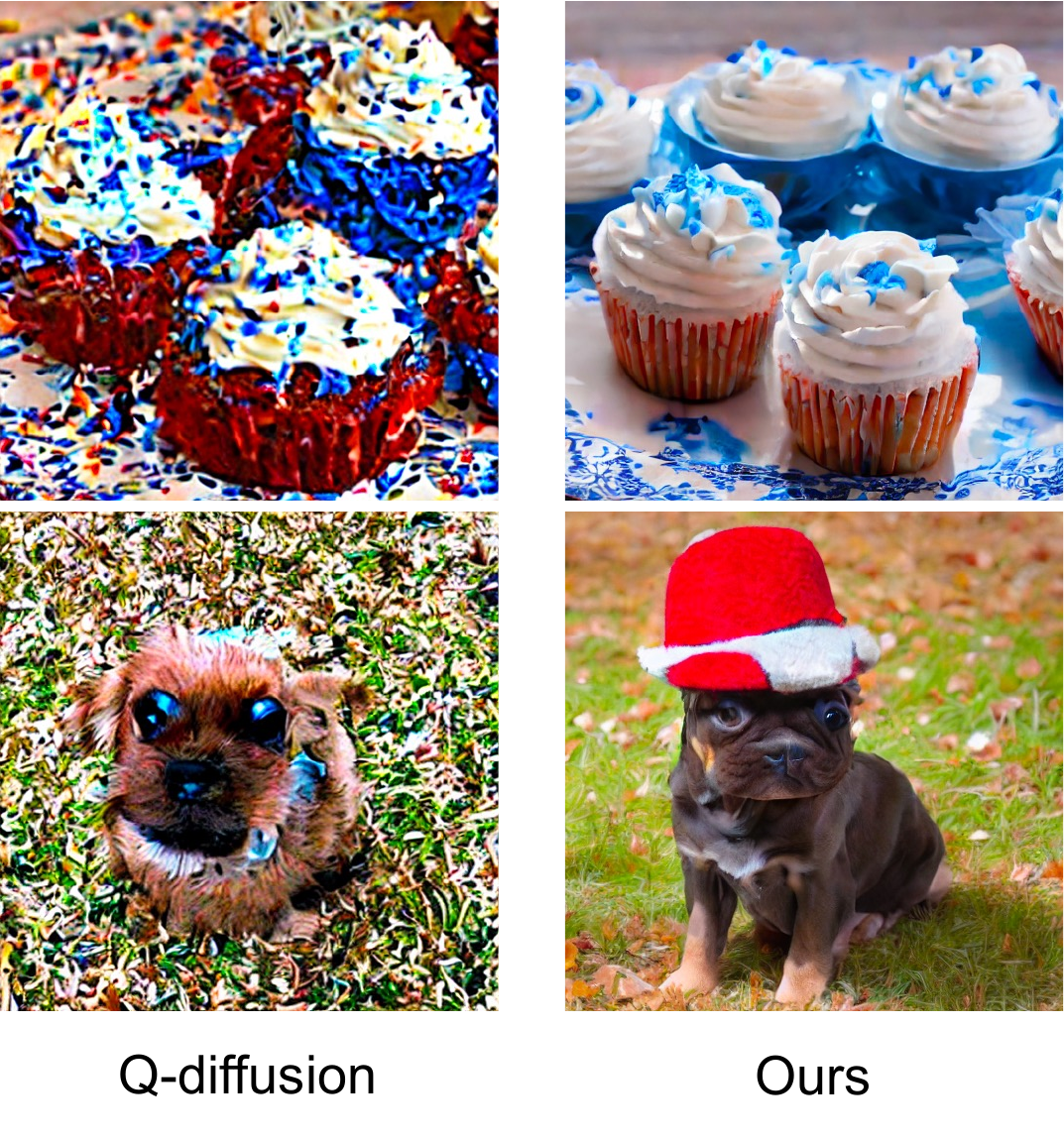}
    \caption{Text-guided generation of Q-diffusion and proposed method with 10 timesteps under BitOPs constraint of W6A6. Input prompts are ``several cupcake with white and blue icing sitting on a plate.''and ``A puppy wearing a hat.''. Compared to Q-diffusion, our method can generate higher quality images.}
    \label{fig:txt2img}
\end{figure}

We conducted our search at 10 timesteps. As shown in Fig.~\ref{fig:txt2img}, our method surpasses the performance of the original W6A6 settings, producing higher-quality images that more closely align with semantic information.

\subsection{Ablation Study and Analysis}
\label{sec:4.3}

\noindent \textbf{Ablation Study. }
\label{sec:4.5}
In this subsection, we examine the impact of our proposed techniques on the CIFAR-10 dataset, using a timestep setting of 10. As indicated in Tab.~\ref{tab:ablation}, implementing a shared mixed-precision quantization across all timesteps is feasible and improves the FID from 13.43 to 9.57. This is reasonable considering the task complexity, which arises from the vast and unassessable space size required for assigning different quantization strategies to various timesteps.
Moreover, by applying non-uniform timestep grouping to minimize the search space, we further enhance performance, reducing the FID from 9.57 to 8.76. Thus, our method demonstrably enhances performance.

\begin{table}[t]
\fontsize{10}{10}\selectfont
  \caption{Ablation studies on CIFAR-10 32$\times$32. ``Shared MPQ policy'' means the mixed-precision quantization policy is identical across each step, ``Non-uniform Grouping'' and ``Uniform Grouping'' are different timestep grouping schemes respectively.} 
  \label{tab:ablation}
  \centering
  \begin{tabular}{@{}cccc@{}}
    \toprule
    Shared MPQ Policy& \makecell[c]{Non-uniform \\ Grouping} & \makecell[c]{Uniform \\ Grouping} & \makecell[c]{FID  ($\downarrow$)} \\
    \midrule
    \ding{53}&\ding{53} & \ding{53} & 13.43\\
    \ding{51}& \ding{53}&\ding{53}  & 9.57 \\

    \ding{51}&\ding{53} & \ding{51} & 9.14\\
    \ding{51}&\ding{51} & \ding{53} & 8.76 \\ 
  \bottomrule
  \end{tabular}
\end{table}

\noindent \textbf{Bitwidth Allocation Distribution.}
As illustrated in Fig.~\ref{fig:bit_width_distribution}, we visualize the bitwidth allocation for weights and activations on the CIFAR-10 dataset for the best candidate at 15 steps. Our method tends to allocate higher bit widths to activations and lower bit widths to weights. This phenomena is rational as aggressive activation quantization may result in significant performance degradation.

\noindent \textbf{Evolutionary Search vs Random Search.}
We assess the effectiveness of evolutionary search in strategy optimization compared to random search. Fig.~\ref{fig:search_process} displays the performance predictions of the ten best strategies identified by evolutionary and random searches over the last seven epochs. The results show that strategies under evolutionary search converge to an optimum around 20, whereas those under random search continue to fluctuate significantly, demonstrating the efficiency of evolutionary search.

\noindent \textbf{Visualization of Timestep Selection.}
Additionally, we visualize the timestep selection policy for CIFAR-10 at 15 steps in Fig.~\ref{fig:exp_comparison}. Here, we plot the cumulative selection counts across timesteps for the top ten optimal selection strategies. Given our method’s non-uniform splitting of timesteps, we use the square root of the timestep index on the x-axis for a clearer presentation. This visualization reveals that most selections concentrate around the midpoint of each timestep group.

\noindent \textbf{Performance on Different Sampling Method.} 
In this section, we conduct experiments on the ImageNet dataset under 6 timesteps to assess our method's enhancements with various sampling methods, including PLMS, DDIM, and DPM-Solver. As presented in Tab.~\ref{tab:different_sampler}, our approach consistently enhances performance irrespective of the sampling method employed. Notably, with PLMS, our method significantly improves the Frechet Inception Distance (FID) from 74.77 to 10.09 and the Inception Score (IS) from 51.11 to 206.13, demonstrating substantial performance gains.

\begin{table}[t]
\fontsize{8}{10}\selectfont
  \caption{FID and IS on ImageNet with different sampler where steps = 6. Our method improves FID and IS performance consistently with DDIM, DPM-Solver and PLMS.}
  \label{tab:different_sampler}
  \centering
  \begin{tabular}{@{}cccc@{}}
    \toprule
     Scheme  & Sampler& FID ($\downarrow$)& IS ($\uparrow$)\\ 
    \midrule
     Q-diffusion & DDIM~\cite{DDIM}& 28.27&135.29 \\
    Ours & DDIM~\cite{DDIM} &\textbf{12.14}&\textbf{229.17}\\\cline{1-4}
     Q-diffusion &DPM-Solver~\cite{dpm-solver}&12.59&211.49 \\
    Ours &DPM-Solver~\cite{dpm-solver}&\textbf{10.50}&\textbf{225.75} \\ \cline{1-4}

     Q-diffusion &PLMS~\cite{PLMS}&74.77&51.11 \\
    Ours & PLMS~\cite{PLMS}&\textbf{10.09}&\textbf{206.13} \\
  \bottomrule
  \end{tabular}
\end{table}

\section{Conclusion}

In this work, we introduce TFMQ-DM, a novel framework designed to jointly optimize timestep selection and quantization precision, thereby enhancing the efficiency of diffusion models. To mitigate the complexities associated with calibrating and assessing various quantization strategies, we develop a mixed-precision solver that facilitates switching between different quantization strategies for performance evaluation. Furthermore, recognizing that different timesteps impact image quality to varying degrees, we propose a non-uniform timestep grouping method that effectively prunes the search space. Our experimental results validate the efficacy of our approach, demonstrating that our method achieves more than $10\times$ BitOPs savings on all tasks while still maintaining equivalent generative performance.

\balance
\bibliography{main}
\bibliographystyle{arxiv}

\end{document}